\documentclass[conference]{IEEEtran}
\IEEEoverridecommandlockouts
\usepackage{cite}
\usepackage{amsmath,amssymb,amsfonts}
\usepackage{algorithmic}
\usepackage{graphicx}
\usepackage{textcomp}
\usepackage{xcolor}
\usepackage{booktabs}
\def\BibTeX{{\rm B\kern-.05em{\sc i\kern-.025em b}\kern-.08em
    T\kern-.1667em\lower.7ex\hbox{E}\kern-.125emX}}
    
\usepackage[export]{adjustbox}
\usepackage{subfig}
\usepackage{listings}
\usepackage[skip=5pt]{caption}
\usepackage{amsmath,amsfonts,bm}

\newcommand{\overbar}[1]{\mkern 1.5mu\overline{\mkern-1.5mu#1\mkern-1.5mu}\mkern 1.5mu}

\newcommand{\fref}[1]{Fig.~\ref{#1}}
\newcommand{\sref}[1]{Sec.~\ref{#1}}
\newcommand{\tref}[1]{Tab.~\ref{#1}}
\newcommand{\eref}[1]{Eq.~\ref{#1}}

\newcommand{\seq}[1]{[\mathbf{#1}]}

\usepackage{wasysym}
\newcommand{\osv}{\emph{OSV}}
\newcommand{\sv}{\emph{SV}}

\newcommand{\seqsetdb}{w}

\newtheorem{definition}{Definition}
\newtheorem{remark}{Remark}
\def\final{} 

\ifdefined\final
\else
\def\enablecomments
\fi

\ifdefined\draft
\def\enablecomments{}
\fi

\usepackage{soul}
\usepackage{xcolor}

\definecolor{LightGreen}{rgb}{0.80,1.00,0.80}
\definecolor{LightBlue}{rgb}{0.80,0.80,1.00}
\definecolor{LightRed}{rgb}{1.00,0.80,0.80}
\definecolor{LightPurple}{rgb}{0.94,0.85,1.00}
\definecolor{LightGray}{rgb}{0.90,0.90,0.90}

\soulregister{\method}{7}
\soulregister{\xspace}{7}
\soulregister{\emph}{7}
\soulregister{\ref}{7}

\ifdefined\enablecomments
  \DeclareRobustCommand{\commentformat}[3]{\sethlcolor{#2}\textsf{\hl{#1: #3}}}
  \newcommand{\sm}    [1]{{\scriptsize\sethlcolor{LightGray}\hl{\textsf{#1}}}}
   
\else
  \newcommand{\commentformat}[3]{}
  \newcommand{\sm}    [1]{}
   
\fi

\newcommand{\caleb} [1]{\commentformat{CL}{LightBlue}{#1}}

\begin{document}

\title{Order-sensitive Shapley Values for Evaluating Conceptual Soundness of NLP Models}

\author{\IEEEauthorblockN{Kaiji Lu}
\IEEEauthorblockA{\textit{ECE Department} \\
\textit{Carnegie Mellon University}\\
Pittsburgh, PA}
\and
\IEEEauthorblockN{ Anupam Datta}
\IEEEauthorblockA{\textit{ECE Department} \\
\textit{Carnegie Mellon University}\\
Pittsburgh, PA}}

\maketitle

\begin{abstract}

Previous works show that deep NLP models are not always conceptually sound: they do not always learn the correct linguistic concepts. Specifically, they can be insensitive to word order. In order to systematically evaluate models for their conceptual soundness with respect to word order, we introduce a new explanation method for sequential data: Order-sensitive Shapley Values (\osv{}). We conduct an extensive empirical evaluation to validate the method and surface how well various deep NLP models learn word order. Using synthetic data, we first show that \osv{} is more faithful in explaining model behavior than gradient-based methods. Second, applying to the HANS dataset, we discover that the BERT-based NLI model uses only the word occurrences without word orders. Although simple data augmentation improves accuracy on HANS, \osv{} shows that the augmented model does not fundamentally improve the model’s learning of order. Third, we discover that not all sentiment analysis models learn negation properly: some fail to capture the correct syntax of the negation construct. Finally, we show that pretrained language models such as BERT may rely on the absolute positions of subject words to learn long-range Subject-Verb Agreement. With each NLP task, we also demonstrate how \osv{} can be leveraged to generate adversarial examples.

\end{abstract}


\section{Introduction}
\label{Introduction}
\sm{previous work: insensitivity to word orders}
Recent works discover that NLP models are not sensitive to word orders in the same way as humans~\cite{pham2020out, sinha2021masked, clouatre2021demystifying, alleman2021syntactic, sinha2021unnatural}: although randomly shuffled sentences are often incomprehensible to humans, they result in very small performance drop for the state-of-the-art Transformer models such as BERT~\cite{devlin2018bert}, for both syntactic and semantic classification tasks.
The underlying hypothesis behind such insensitivity is that the occurrences or co-occurrences of words are often sufficient for learning the tasks~\cite{malkin2021studying}. However, relying on such heuristics without word order is not \emph{conceptually sound}~\cite{parkinson2011sr}, i.e. the models fail to learn the correct linguistic concepts. One aspect of conceptual soundness, which is the focus of this paper, is \emph{order-sensitivity}. A model may ignore order-sensitive concepts and thus become susceptible to adversarial examples. One notable example is HANS~\cite{mccoy2019right}, which discovers that Natural Language Inference~(NLI) models wrongly predict entailment between hypothesis and premise sentences with overlapping words but different ordering, such as \textit{The doctors visited the lawyers} entailing \textit{The lawyers visited the doctors}. 
\begin{figure}[t]
\centering
\includegraphics[width=\columnwidth]{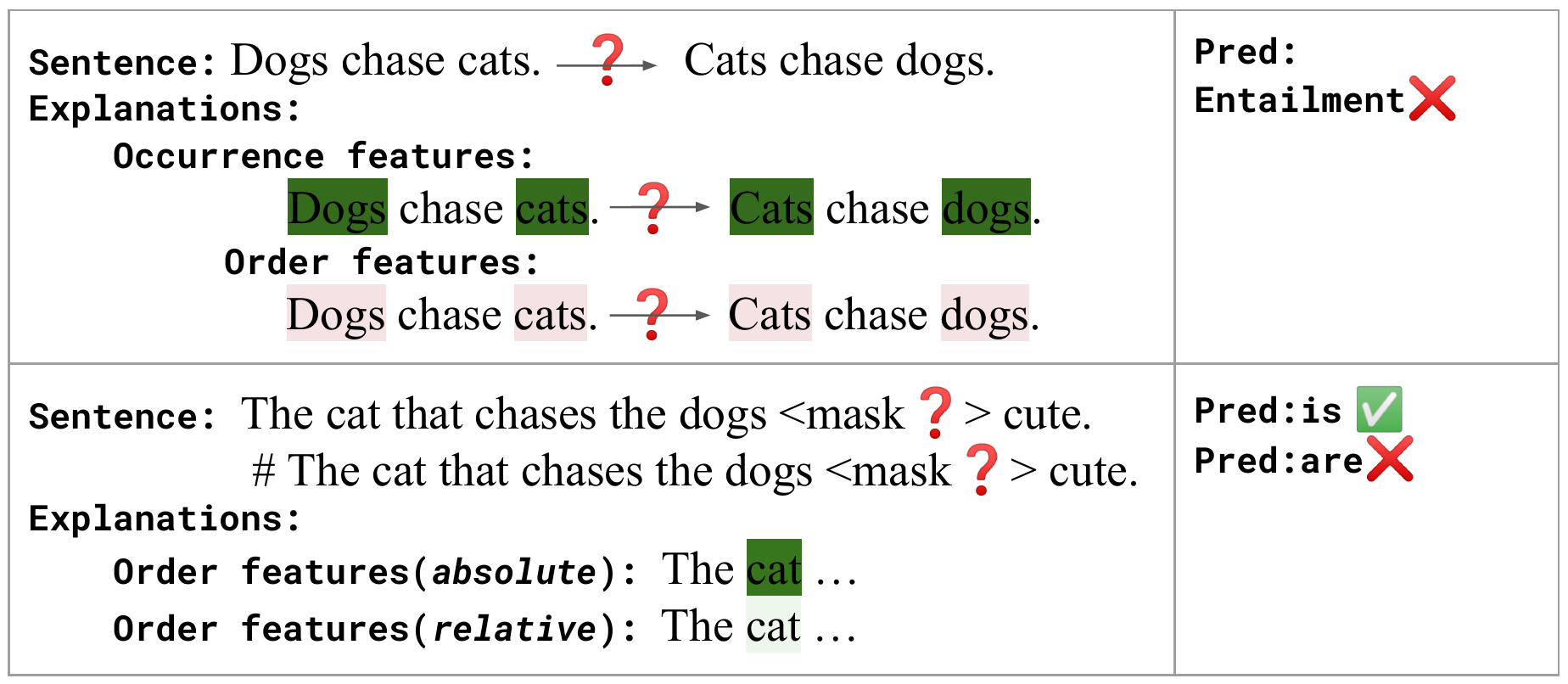}
\caption{Two examples of Order-sensitive explanations. Two colors represent opposite signs of attributions while darker shade represents larger magnitude. Top: In an NLI model, attributions to the entailment class are small and negative for order features (light pink) but large and positive for occurrence features (dark green), resulting the wrong prediction of entailment between inverted sentences; Bottom: To learn subject verb agreement: a language model relies on absolute positions of the subject word (larger attribution to absolute order features than relative ones), resulting in an error when the subject's position is shifted by prepending the token \#. }
\label{fig:intro}
\end{figure}
\sm{previous work: order-insensitive shapley values}
To systemically understand how and to what extent the models capture order information, a method specially designed to explain sequence orders is needed.  

Game theoretic approaches such as Shapley values~\cite{shapley1953contributions} are widely adopted to explain the predictions of deep models by assigning influences to features~\cite{strumbelj2010efficient, datta2016algorithmic,lundberg2017unified, sundararajan2020many, covert2020understanding}. Prior work applying Shapley values to textual data or other sequential data, however, treat sequential data as tabular data without attributing to the order of features in a sequence. 

\sm{how OSSV explanations can be used to evaluate and diagnose the conceptual soundness of a model}
In this work, we propose a new model-agnostic explanation framework, \emph{Order-sensitive Shapley Values}~(\osv{}), for attributing to the \emph{order} of features along with \emph{occurrence} of features in a sequence. We propose a mechanism for separating these two groups of features and two modes of intervening on sequence orders, \emph{absolute position} and \emph{relative order}. Most importantly, we show that \osv{} as an explanation device can help answer the questions of \emph{conceptual soundness} of NLP models: \textbf{does the model learn the correct order-sensitive syntax, or is the model exploiting mere occurrences of words that are possibly spurious correlations? }
We conduct an extensive empirical study applying \osv{} to synthetic datasets and several NLP tasks: Natural Language Inference~(NLI), Sentiment Analysis~(SA) and Subject Verb Agreement~(SVA). For each task, we demonstrate how \osv{} is the appropriate explanation device to precisely diagnose the root cause of spurious correlations by constructing adversarial examples based on the generated explanations. Two examples of \osv{} applied to NLI and SVA are shown in \fref{fig:intro}. 
 Our contributions and findings are summarized below:

\begin{itemize}
    \item We propose Order-sensitive Shapley Values (\osv{}) by expanding the standard feature set to include  \emph{order features} (\sref{sec:method}). Using synthetic data, we show that \osv{} is more faithful in explaining model's behavior than gradient-based methods (\sref{sec: toy});

    \item Applying \osv{} to the HANS dataset~\cite{mccoy2019right}, we discover that BERT uses only word occurrences without word orders. 
    Data augmentation~\cite{min2020syntactic} improves accuracy on HANS, but \osv{} shows that the augmented model does not fundamentally improve the model's learning of order features (\sref{sec: hans});
    \item We discover that sentiment analysis models, such as BERT, do not learn negation properly: they fail to capture the correct syntax of negators preceding adjectives. Interestingly, RoBERTa~\cite{liu2019roberta} and StructBERT~\cite{wang2019structbert} produce more conceptually sound explanations than BERT~\cite{devlin2018bert}(\sref{sec: sst});
    \item We show that pretrained language models such as BERT rely on the absolute positions of subject words for capturing long-range Subject-Verb Agreement~(\sref{sec: sva}).
    \item In each NLP task, we show how insights drawn from \osv{} can be exploited to generate adversarial examples. 

\end{itemize}
\section{Background}
\subsection{Shapley value as an attribution method}\label{sec: bg-shap-attr}
A large group of explanation methods, summarized by~\cite{covert2021explaining}, quantify the impact of individual features (or groups of features) by measuring how removing those features change the outcome of model predictions. 
In this paper, we focus on one axiomatic method inspired by coalition game theory, Shapley values~\cite{shapley1953contributions}. Given a set of $n$ players represented by integers $N = \{0,1,\dotsc,n-1\}$ and a value function $v:2^n\mapsto \mathbb{R}$, the \emph{Shapley value} of player $i$ is:
\begin{equation}
    \phi_{v}(i)=\sum_{S \subseteq N \backslash\{i\}} \frac{|S| !(n-|S|-1) !}{n !}\Big(v(S \cup\{i\})-v(S)\Big)\label{eq:shapley}
\end{equation}



In the machine learning setting, to apply Shapley values for a classification model $f$ with input features $x = \{x_0,x_1, \dotsc, x_{n-1}\}$, $x$ are treated as individual players and  a function $f^y(x):X\mapsto \mathbb{R}$ related to the model output $f(x)$ is treated as the value function. For example, $f^y(x)$ can be the output probability of the predicted class for local explanations~\cite{lundberg2017unified}, a loss function for global explanations~\cite{covert2020understanding}, or any user-defined function of the output logits~\cite{datta2016algorithmic}. 

\paragraph{Intervention on removed features} A distribution $g(x_{\overbar{S}}|x_S)$ for intervening on removed features~($\overbar{S} = N\backslash S$) is required to compute $f^y(x)$ for feature subset $x_S$,  The choice of $g$, however, has been a subject of debate in recent studies~\cite{chen2020true, kumar2020problems, catav2021marginal, frye2020shapley, hase2021out}.~\cite{chen2020true} in particular describes a choice between either conditional or interventional expectation for $g$: the former respects the joint distributions of features to explain the \emph{data}, while the latter breaks the correlation among features to explain the \emph{model}. In this paper, we adopt the latter: specifically, we inspect model's response to reordered sequences of words that may deviate from the underlying data distribution, allowing for a thorough understanding of the model's mechanism of encoding orders in sequences. 


\sm{some argument on the current of applying Shapley values to sequences}
\section{Method}
In this section, we present the extension of Shapley Values~(\sv{}) into Order-sensitive Shapley Values~(\osv{}) by expanding the feature set with \emph{order features}. We explore the connections between the two and propose two mechanisms of intervention on order features, attributing to the \emph{absolute} or \emph{relative} positions of elements in a sequence, respectively. 

\subsection{Set Representation of a Sequence}\label{sec:method} \sm{defining a sequence}


We start by defining a sequence as a \emph{set} of features. A finite ordered sequence $\seq{w}$ can be represented by a union of two sets  $\seqsetdb = x \cup z$ where the \emph{occurrence feature} set $x = \{x_0,x_1, \dotsc, x_{n-1}\}$ and the integer \emph{order feature} set $z = \{z_0,z_1, \dotsc, z_{n-1}\} = N$, with a bijective mapping  $h:z\mapsto x$,  $h(z_i) = x_i, \forall i\in N$. The sequence is simply:
\begin{gather}\label{eq:wseq}
    \seq{w}_{z_i} = x_i, \forall i\in N \notag
\end{gather}
We use $\mathbf{w}$, $\mathbf{x}$ and $\mathbf{z}$ to represent sets with ordered assignment of values for elements in otherwise orderless sets $w$, $x$ and $z$: $\mathbf{w} = (x_0,x_1, \dotsc, x_{n-1},z_0,z_1, \dotsc, z_{n-1})$ and similarly for $\mathbf{x}$ and $\mathbf{z}$. [$\mathbf{w}]$ can also be represented from $w$ by enumerating elements of $x$ indexed by the corresponding elements in $z$:
$\seq{w} = \pi_{\mathbf{z}}(\mathbf{x})$.
In the opposite direction of the mapping, since many combinations of $x$ and $z$ represent the same sequence $\seq{w}$, we assume by default $\mathbf{z} = (0,1,\dotsc,n-1) = \mathbf{N}$ when mapping $\mathbf{w}$ to $w$. We use $\mathbf{w}$, $w$ and $\seq{w}$ interchangeably when referring to the same sequence. 
\subsection{Order-sensitive Shapley Value for Sequential Inputs}
A sequence $\seq{w}$ of length $n$ can be treated as a game with $2n$ players with the feature set $\seqsetdb$. Attributions to $w$ measure how important it is for a feature to be present, while attributions to $z$ measure how important it is for a feature to be in the correct position, they are computed together by plugging in \eref{eq:shapley}:
\begin{definition}[Order-sensitive Shapley Value]
\begin{align}
    \phi^z_{f^y}(w_i)&=\sum_{S \subseteq N' \backslash i} \frac{|S| !(2n-|S|-1) !}{(2n)!}m(S) \label{eq:oshapley}
\end{align}
\begin{align}
    \text{with }
    m(S) &= f^y( \mathbf{w}_{S \cup i})-f^y( \mathbf{w}_S) \notag\\
    f^y(\seqsetdb_S) &= \mathbb{E}[f^y(\mathbf{w}) | \mathbf{w}_S] \notag \\
                &= \mathbb{E}_{\mathbf{w}_{\overbar{S}}| \mathbf{w}_{S} }[f^y(\mathbf{w}_S)] \notag \\ 
                &\approx \mathbb{E}_{\mathbf{z}_{\overbar{S_z}}|\mathbf{z}_{S_z}} \mathbb{E}_{\mathbf{x}_{\overbar{S_x}}|\mathbf{x}_{S_x}\cup \mathbf{z}_{S_z}}[f^y(\mathbf{x}\cup \mathbf{z}_{S_z})] \label{eq:fy_xz}  \\
                &\approx \mathbb{E}_{\mathbf{z}\sim q (\cdot|\mathbf{z}_{S_z})}   \mathbb{E}_{\mathbf{x}\sim g (\cdot | \mathbf{x}_{S_x}\cup \mathbf{z})} [f^y(\pi_\mathbf{z}(\mathbf{x}))] \label{eq:fy_pi} 
\end{align}
\begin{align}
   \text{where: } N' &= \{0,1,\dotsc,2n-1\} \notag \\
   S_x &= \{i|\mathbf{w}_i \in x \land i \in S\}, \overbar{S_x} = N \backslash S_x \notag\\ 
   S_z &= \{i-n|\mathbf{w}_i \in z\ \land i \in S\}, \overbar{S_z} = N \backslash S_z\notag
\end{align}
\end{definition}
$S\cup i$ abbreviates $S\cup \{i\}$. Choices of $g$ is discussed in \sref{sec: bg-shap-attr} and choices of $q$ is to be discussed in \sref{sec: replacing}. In \eref{eq:fy_xz}, we assume features in $z$ are independent from those in $x$, in other words, how occurrence features can be ordered into a sequence does not depend on the value of the occurrence features themselves. 
In \eref{eq:fy_pi}, we assume that the intervention of occurrence features is only done in the context of the original sequence. 
With these two assumptions, we align \osv{} with \sv{}
through the following remarks:



\begin{remark}\label{rem: remark1}
when $q(\cdot) = \mathbf{N}$, $\phi^z$ reduces to the  (order-insensitive) Shapley values $\phi$  as follows:
$$\phi^z_{f^y}(x_i) = \phi_{f^y}(x_i), \forall x_i \in x $$
$$\phi^z_{f^y}(z_i) = 0, \forall z_i \in z$$
\end{remark}
The proof is included in Appendix~\ref{appendix:method}. Remark~\ref{rem: remark1} states that if we assume the omnipresence of order features, \osv{} for order features are all zero and \osv{} for occurrence features reduce to corresponding order-insensitive Shapley values~(\sv{}). 
\caleb{Proof in appendix}
\begin{remark}\label{rem: remark2}
If $X$ is  the set of intervened sequences used for computing \sv{}, and $Z$ for computing the corresponding \osv{}, then $Z \subseteq \{z | z \in \Pi(x), \forall x \in X\}$. where $\Pi$ is a function mapping to all permutations of elements in $X$: $\Pi(\mathbf{w}) = \{\pi_\mathbf{z}(\mathbf{w}): \forall \mathbf{z}\}$.
\end{remark}
Remark \ref{rem: remark2} stipulates that the space of intervened sequences evaluated for $\phi^z$ is encompassed by permutating those evaluated for the corresponding $\phi$. The intervention on order features is therefore orthogonal to the choice of $g$, enabling the potential expansion of order features to other framework besides Shapley values, such as Banzhaf values~\cite{banzhaf1964weighted} or LIME~\cite{ribeiro2016should}. 

\begin{remark}\label{rem: remark3}
A model $f^y$ is \emph{completely order-sensitive} if $f^y(\mathbf{w}_S) = y_{\varnothing}, \forall S_z \neq N $: 
\begin{align}
  \phi^z_{f^y}(x_i) &= \sum_{S_x \subseteq N \backslash i} \frac{(|S_x| + n)!(n-|S_x|-1) !}{(2n)!}m(S_x) \\
  &= \sum_{S_x \subseteq N \backslash i} p(S_x) \frac{|S_x| !(n-|S_x|-1) !}{n!}m(S_x) \label{eq: reweight}\\
  \text{where: } p(S_x) &=\frac{(2n - |S_x| - 1)!n!}{(n - |S_x| - 1)!(2n)!} \notag\\ 
\forall z_i , \phi^z_{f^y}(z_i) &= \sum_{S_x \subseteq N} \frac{|S_x + n| !(n-|S_x|-1) !}{(2n)!}[f^y( \mathbf{x}_{S_x})-y_{\varnothing})] \notag 
\end{align}
\end{remark}
Remark \ref{rem: remark3} describes a hypothetical, \emph{completely order-sensitive model} $f^y$ undefined on all reordered sequences~(those that miss any feature from $z$), thus output a constant ($y_{\varnothing})$). For example, the joint distribution of words in a sentence for most natural languages is so sparse that almost all reordered sentences are out of distribution. \eref{eq: reweight} shows the relation between the attribution to $\phi^z$ and the corresponding $\phi$: each term in the sum of \eref{eq:shapley} is reweighted by $p(S_x)$. Attributions to order features ($\phi(z)$) are all equal since missing any order feature results in $y_{\varnothing}$. For longer sequences, $p(S_x)$ becomes extremely small, making $\phi(x)$ negligible compared to $\phi(z)$. In other words, under a \emph{completely order-sensitive model} $f^y$, \osv{} should be dominated by order features. However, in the real world, neither a human nor an NLP model is completely order-sensitive. Some tasks inherently tolerate reordered sentences than others: For a sentiment analysis model, for instance, even though \textit{film the good very is} is technically not a valid sentence thus no sentiment shall be assigned, most models~(and human) would still justifiably label it with positive sentiment, while more syntactic tasks such as language modeling should be far more order-sensitive, where model's response to incoherent sentences should be more neutral therefore $\phi(z)$ should be larger. This hypothesis is confirmed by various NLP tasks evaluated in \sref{sec:nlp}.

\subsection{Intervention on order features}\label{sec: replacing}
Assuming the independence of order features from occurrence features, the only missing piece left is to choose an appropriate $q$ for sampling $\mathbf{z}\sim q (\cdot|\mathbf{z}_{S_z})$.
A key premise is that any $q$ should guarantee a valid sequence, i.e,  $q (\cdot|\mathbf{z}_{S_z})= N $. We propose the following two $q$, evaluating two distinctive notions of ordering: \emph{absolute position} and \emph{relative order}. 
\begin{definition}[Absolute-position order intervention]\label{def:absolute position}
$$q^a(\cdot|\mathbf{z}_{S_z})  = \text{Unif}(\Pi(\mathbf{z}_{\overbar{S_z}}); \mathbf{z}_{S_z})$$
\end{definition}
For example, for a sequence $\seq{w} = (a,b,c,d)$ and $S_x = \{0,1,2,3\}$, $S_z = \{0,1\}$, then $q^a(\cdot|\mathbf{z}_{S_z}) = \text{Unif}\{(a,b,c,d), (a,b,d,c)\}$.
\begin{definition}[Relative-order order intervention]\label{def:relative order}
$$q^r(\cdot|\mathbf{z}_{S_z})  = \text{Unif}(\Pi(\mathbf{z}_{\overbar{r(S_z)}}); \mathbf{z}_{r(S_z)})$$ where $r(\cdot)$ does a random non-wrapping shifting of the feature positions for unintervened features.  
\end{definition}
With the setting of the previous example, $q^r(\cdot|\mathbf{z}_{S_z}) =\text{Unif}\{\allowbreak((a,b,c,d), \allowbreak(c,a,b,d), \allowbreak(c,d,a,b), \allowbreak(a,b,d,c),\allowbreak (d,a,b,c), \allowbreak(d,c,a,b)\}$.

\sm{difference between the two}
We denote \osv{} computed with $q^a$ and $q^r$ as $\phi^a$ and  $\phi^r$, respectively.  It is apparent from Def.\ref{def:absolute position} and \ref{def:relative order} that $Supp(q^a) \subseteq Supp(q^r)$.  With $\phi^a$, we consider an order feature as important if deviating from its absolute position cause a large change in $f^y(x)$. With $\phi^r$, on the other hand, an order feature is only considered as important if breaking its relative ordering with other features affects $f^y(x)$. For natural language or other variable-length data, $\phi^r$ can help assess if the model captures the order of $n$-gram features. Similar notions of shuffling $n$-gram features have been explored in~\cite{sinha2021masked} and~\cite{alleman2021syntactic}, while $q^r$ summarize all $n$-gram coalitions and attribute them to individual order features. 
$q^a$, on the other hand, follows a stricter and more general notion of \emph{order} where the position of unintervened order features will not change. $q^a$ may also work well with sequential data with a meaningful starting position such as time-series data. The difference between $q^a$ and $q^r$ also functions as a diagnostic tool to see if a model relies on the absolute position of features when it should not: \sref{sec: sva} explore an NLP example on SVA. 

Note that all axioms of Shapley values~\cite{shapley1953contributions, datta2016algorithmic} extend to \osv{}. For example, the completeness axiom states:
$\sum_i^{N'} \phi^z_{f^y}(i) = f^y(\mathbf{w}) - f^y(\mathbf{w_\varnothing})$, i.e., all feature attributions~(both occurrence and order features) sum up to the difference between the prediction of the instance and that of an empty baseline. The complete set of axioms and how they can be interpreted for order features is included in Appendix~\ref{appendix:method}. 

One disadvantage of Shapley value is that it is often computationally intractable, which could be exacerbated by the doubling of feature size for \osv{}. However, many approaches have since been proposed to address this issue such as SampleShapley~\cite{strumbelj2010efficient, datta2016algorithmic}, SHAP~\cite{lundberg2017unified}, and DASP~\cite{ancona2019explaining}, all of which are extendable to \osv{}. In \sref{sec:eval}, we use \osv{} as global explanations~\cite{covert2020understanding} for faster approximation, especially when we are only interested in overall feature importance. Please refer to Appendix~\ref{appendix:impl} for more details.




\section{Evaluation}\label{sec:eval}
In this section we apply \osv{} to a variety of tasks for both synthetic data~(\sref{sec: toy}) and natural language tasks~(\sref{sec:nlp}). Specially, we show how \osv{} offers richer and deeper insights in explaining models' behavior compared to prior methods such as \sv{} and gradient-based explanation methods, and how order-insensitivity shown by \osv{} precisely uncovers model's weakness to potential adversarial manipulations. 
\subsection{Synthetic Data Experiment}\sm{1 page} \label{sec: toy}
Since it is often difficult to evaluate the faithfulness~\cite{jacovi2020towards} of an explanation method~\cite{ju2021logic, zhou2021feature}, we first experiment with synthetic data where ground truth explanations are available. Our data and models are inspired by a similar experiment by~\cite{lovering2020predicting}.

\paragraph{Models and Data}
 We experiment with two model architectures: (1)~an RNN with 4-layer LSTM~\cite{hochreiter1997long} and an MLP with 1 hidden layer and ReLU activation. The embedding dimension $E$ and hidden size $H$ of LSTM are both 512. (2)~a Transformer encoder model~\cite{Vaswani2017AttentionIA} with 2 self-attention layers, 8 attention heads and same $E$, $H$ and MLP layer as the RNN model. We train 5 randomly seeded models for three $k$-length binary sequence classification tasks, with a symbolic vocabulary $V$ of size 200 containing integer symbols $(0 \cdots |V|-1)$ and $k=8$. The datasets each contain 400k sequences with a 99/1 split between training and testing. The description of the tasks and performance of the LSTM models are included in \tref{tab: toy-desc} and those of the Transformer models are in Appendix~\ref{appendix:toy}. All models have a test accuracy of at least 98\%, ensuring that the models truly generalize to the tasks. 
 
  For each model, we compute global Shapley explanations~\cite{covert2020understanding}, both order-insensitive($\phi$) and order-sensitive($\phi^a$, $\phi^r$), with a uniform discrete distribution over all symbols for $g$. We use 5 random seeds for each explanation to control for the randomness from $g$ and $q$ (total of 25 random seeds per task). We compute $\phi$, $\phi^a$ and $\phi^r$ on 1000 1-labeled sequences of the test-set of Task 1, $\seq{W_1}$. $\seq{W_1}$ satisfies the condition of ``sequence beginning with a duplicate''(ex. $[1 1 2 3 4 5 6 7]$).  We choose $f^y(x)$ as the difference between the predicted probability of the correct class~($y=1$) and that of the incorrect class~($y=0$), as used in previous works~\cite{leino2018influence, datta2016algorithmic}. Similar $f^y(x)$ is used in the rest of the paper. Since the condition of Task 1 is sufficient for that of Task 2 and Task 3, Model 2 and Model 3 can also accurately classify $\seq{W_1}$, as shown by \textit{Acc-1} of \tref{tab: toy-desc}. 
  
 \begin{table}[t]
\caption{Tasks and performance of LSTM models. Acc.: mean test accuracy for each model over 5 random seeds, Acc-1: mean test accuracy for $\seq{W_1}$.}
\label{tab: toy-desc}
\vskip 0.15in
\begin{small}

\begin{center}
\begin{tabular}{lcccr}
\toprule
Ind. & Condition for y=1 & Acc. & Acc-1. \\
\midrule
1    & Sequence beginning with duplicate & 1.00$\pm$ 0.00& 1.00$\pm$ 0.00\\
2   & Adjacent duplicate in the sequence  &1.00$\pm$ 0.00 &1.00$\pm$ 0.00 \\
3    & Any duplicate in the sequence &0.99$\pm$ 0.00& 1.00$\pm$ 0.00 \\

\bottomrule
\end{tabular}
\end{center}
\end{small}
\vskip -0.1in
\end{table}

\paragraph{Result}
\begin{figure}[t!]
\centering
\subfloat[Model 1: Begins with duplicate]{\label{fig:toy_lstm_1}
     \includegraphics[width=.8\columnwidth,valign=c]{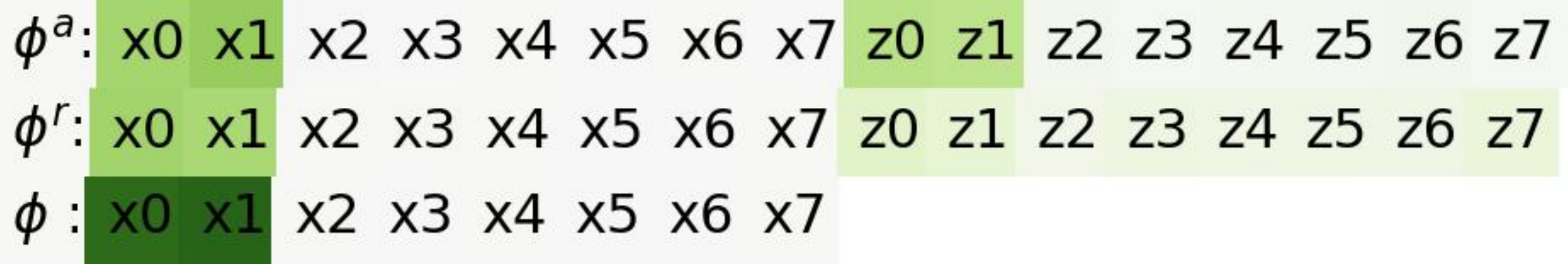}}

\subfloat[Model 2: Adjacent duplicate]{\label{fig:toy_lstm_2}
     \includegraphics[width=.8\columnwidth,valign=c]{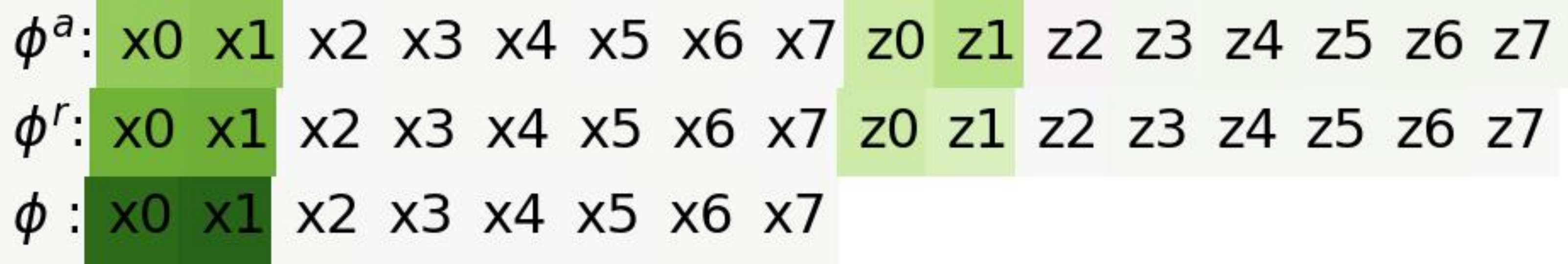}}

\subfloat[Model 3: Any duplicate]{\label{fig:toy_lstm_3}
     \includegraphics[width=.8\columnwidth,valign=c]{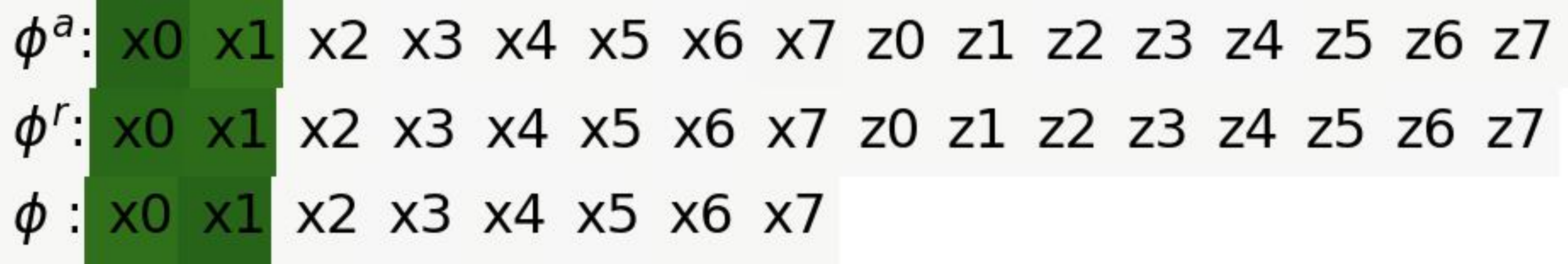}}
\caption{Explanations($\phi^a$, $\phi^r$ and $\phi$) of $\seq{W_1}$ by three LSTM models solving tasks in \tref{tab: toy-desc}. Positive explanations are green and Negative explanations are pink. Darker shade represents larger magnitude.}
\label{fig:toy_lstm}
\end{figure}

\fref{fig:toy_lstm} shows explanations of $\seq{W_1}$, for three models, respectively. Applying \sv{} which only attributes to occurrence features $x$, we obtain identical explanations across three models: $x_0$ and $x_1$ are equally influential while other features have zero attribution. \osv{}, on the other hand, faithfully recovers the importance of sequence order by also attributing to index features $z_0$ and $z_1$ for Model 1 and 2, but not for Model 3.
 
Varying $q$, we observe that $q^r$ for Model 1 does not attribute exclusively to $z_0$ and $z_1$ as $q^a$ does: $z_0$ and $z_1$ does not function as a 2-gram feature, as they are only influential in their respective absolute positions. Instead, the attributions distribute evenly to all order features, as intervening on any order feature almost always cause $\seq{w}_0$ and $\seq{w}_1$ to change positions. On the other hand, $\phi^r$ and $\phi^a$ are similar for Model 2, as $\seq{w}_0$ and $\seq{w}_1$ can be treated as a 2-gram feature, which is influential as long as their relative position is retained.  The results for Transformer models are highly similar to \tref{tab: toy-desc} and \fref{fig:toy_lstm}, which is included in the Appendix \ref{appendix:toy}.
 
 \paragraph{Comparison with baselines} 
 Since our paper is the first to define attributions for order features, there are no model-agnostic baseline methods to directly compare with \osv{}. However, Transformer models separately encode orders with positional embeddings and occurrences of words with word/token embeddings~\cite{Vaswani2017AttentionIA}, analogous to order features and occurrence features. As a result, we compare \osv{} with three baselines across two gradient-based methods, Saliency map~\cite{simonyan13saliency} and Integrated Gradients (IG)~\cite{sundararajan2017axiomatic} by attributing to two embeddings separately. For IG, we use the zero baseline for the word embeddings and experiment with two baselines for position embeddings: (1) zero embeddings ($\phi_{IG}^0$); (2) a dynamic baseline ($\phi_{IG}^p$) which randomly permutes the position ids used to create position embeddings (equivalent to shuffling a sequence), and iterate through all permutations or until it converges. 

To compare the explanations with the ground truth (Ex. $[1,1,\allowbreak 0,0,\allowbreak0,0,\allowbreak0,0]$ for both $\phi^a(x)$ and $\phi^a(z)$ for Model 1), we compute two Pearson correlations (1) $p_a$ between $\phi$, $\phi^r$, $\phi^a$ and the ground truth(16 features). We set the missing $\phi(z)$ values of $\phi$ to be zero. (2) $p$ between $\phi$s and the ground truth, for the occurrence features($\phi(x)$) alone(8 features).  As shown in \tref{tab: toy-quan}, explanations using \osv{} are more accurate to recover the order-sensitive ground truth~($p_a$) for both LSTM and Transformer models, while retaining the same accuracy for occurrence features~($p$). Neither $\phi_{IG}^0$ and $\phi_{IG}^p$ recovers attributions to both features or even to the occurrence features.  

There are three potential reasons for the poor faithfulness of gradient-based methods, especially IG, for attributing to position embeddings: (1) Intervention using gradients such as $\phi_{saliency}$ is not a faithful intervention for ``removing order features''; (2) Despite axiomatic properties of IG, a fixed baseline for positional embeddings result in invalid sequences while the dynamically permutated baseline does not easily converge; \caleb{appendix} (3) Both IG baselines intervene continuously on the sparse positional embeddings space, resulting in potentially invalid sequence representations. Most importantly, not all models explicitly encode order features as positional embeddings, making IG fundamentally inapplicable for non-Transformer models such as RNN. In comparison, \osv{} is model agnostic.

\begin{table}[t]
\caption{Correlations with the ground truth for various explanation methods.}
\label{tab: toy-quan}
\begin{small}
\begin{center}
\begin{tabular}{lll|ll}
\toprule
&  \multicolumn{2}{l|}{LSTM} & \multicolumn{2}{l}{Transformer} \\
& $p_a$  & $p$ & $p_a$  & $p$ \\
\midrule
$\phi^a$  & 0.97$\pm$0.02&0.99$\pm$0.01 & 0.96$\pm$0.03&0.99$\pm$0.01  \\
$\phi^r$ &  0.90$\pm$0.08&0.99$\pm$0.00 & 0.89$\pm$0.08&0.99$\pm$0.00 \\
$\phi$  & 0.76$\pm$0.16&0.99$\pm$0.01 & 0.76$\pm$0.16&0.99$\pm$0.01 \\
$\phi_{saliency}$    & -& - &  0.40$\pm$0.54&0.35$\pm$0.61  \\
$\phi_{IG}^0$    & -& - &  0.05$\pm$0.36&0.55$\pm$0.55  \\
$\phi_{IG}^p$ & -& - &  0.12$\pm$0.27&0.52$\pm$0.61 \\

\bottomrule
\end{tabular}
\end{center}
\end{small}
\vskip -0.1in
\end{table}



\subsection{Natural Language Experiment}\label{sec:nlp}

\subsubsection{Hans}\label{sec: hans}
\paragraph{Models and Data}
HANS Challenge set~\cite{mccoy2019right} is constructed to test the syntactic understanding of NLI models against spurious heuristics of overlapping words between the premise and the hypothesis sentences. For example, the sentence \textit{The doctors visited the lawyers} should not entail \textit{The lawyers visited the doctors} even though they contain the same words. While BERT fine-tuned on the MNLI corpus~\cite{williams2018broad} achieves high accuracy on its test set, its accuracy on the non-entailment subset of HANS is less than $20\%$. To understand the role of word order play in a task such as NLI, we apply \osv{} to the original BERT model fine-tuned on MNLI, $M_{ori}$ and two models designed to be robust against HANS dataset:
\begin{itemize}
    \item $M_{aug}$ using data augmentation.~\cite{min2020syntactic} augments the training dataset of MNLI with instances in the form of `` $A$ $\not\to$ inverted $A$'', such as \textit{This small collection contains 16 El Grecos} $\not\to$ \textit{16 El Grecos contain this small collection}, where $A$ are the hypothesis sentences of the original MNLI corpus. $M_{aug}$ scores $67\%$ on HANS. 
    \item $M_{for}$ using BoW Forgettables.~\cite{yaghoobzadeh2021increasing} retrains BERT on instances unlearnable by a simple Bag-of-Words model. It enables the model to focus on ``forgettable examples'' which interestingly contains instances with the same heuristics as those targeted by HANS. $M_{for}$ scores $72\%$ on HANS. 
\end{itemize}

Both models have exactly the same architecture~(BERT-Base~\cite{devlin2018bert}) as $M_{ori}$, and neither makes use of secondary datasets including HANS itself, making their explanations more comparable. Similar to \sref{sec: toy}, we compute global explanations for all sentences of a same template, for all templates used to construct HANS, and intervene with a discrete uniform distribution of words at each template position. 
\begin{figure}[t]
\centering
\includegraphics[width=\columnwidth]{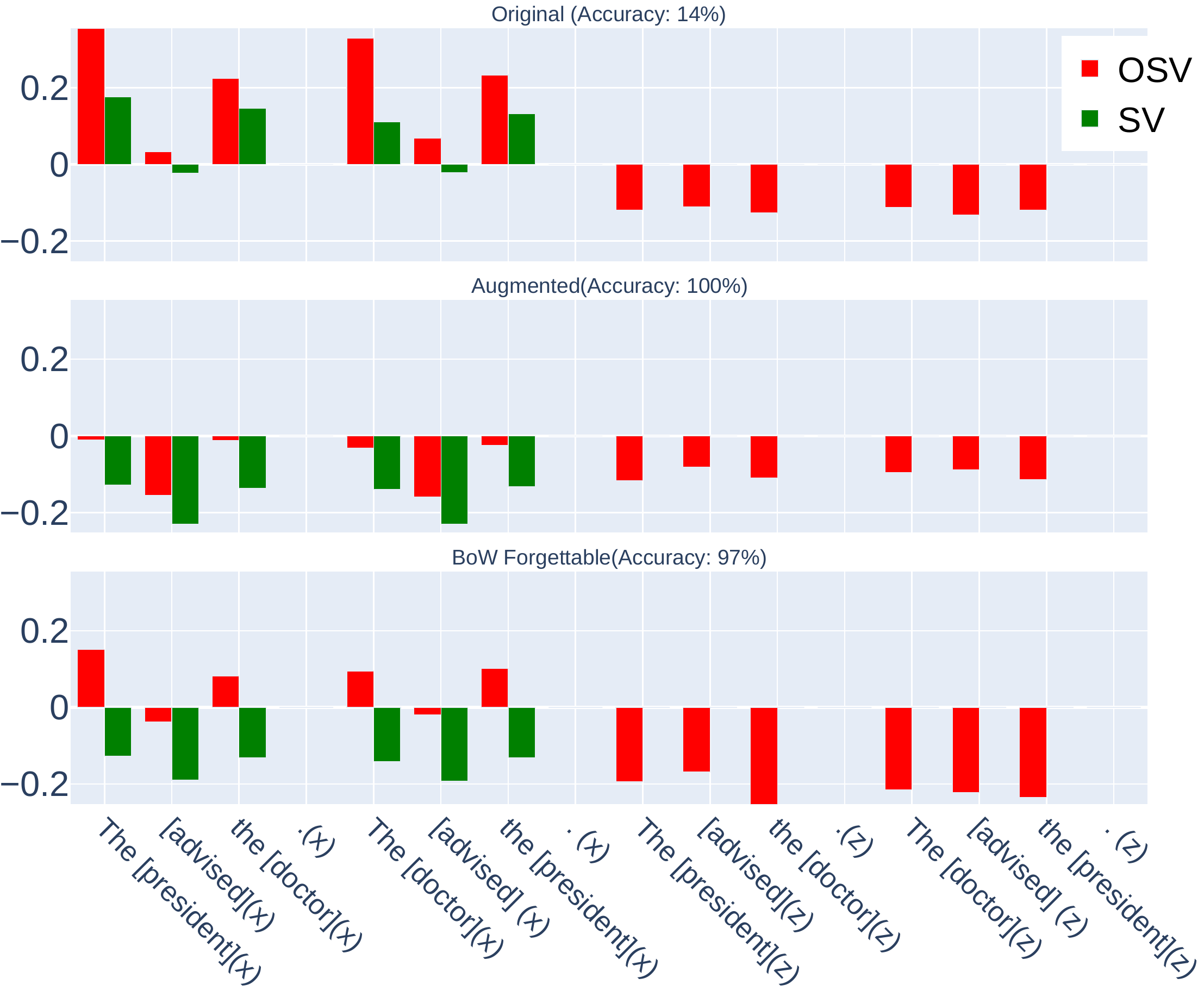}
\caption{Global explanations for instances with a template of HANS where nouns are swapped between the premise and the hypothesis, and the ground truth is non-entailment. Features suffixed by (x) and (z) are occurrence features and order features, respectively. Words in brackets ([]) are one example filling in variable template positions.  Attributions are computed for $f^y(x) = p_{entailment} -p_{non-entailment}$, so positive attribution values indicate the features drive the model towards entailment while negative values indicates the features drive the model towards non-entailment. For easier visualization, attributions of \emph{the} are combined with that of the following nouns.}
\label{fig:mnli_temp1}
\end{figure}



\begin{table}[t]
\caption{Global Statistics across all templates and accuracy on HANS \& HANS*}
\label{tab: hans}
\begin{small}
\begin{center}
\begin{tabular}{lcccc}

\toprule
 & $\sum_i\phi(x_i)$   & $\sum_i\phi(z_i)$  & Acc-HANS & Acc-HANS*  \\
\midrule
$M_{ori}$ & 1.01 &-0.60 & 0.58  &1.00\\
$M_{aug}$  &0.36($\downarrow$0.65)  &-0.58($\uparrow$0.02)  & 0.67 &0.64\\
$M_{for}$  &0.37($\downarrow$0.64)  &-0.69($\downarrow$0.09)  &0.72 &0.94\\

\bottomrule
\end{tabular}
\end{center}
\end{small}
\end{table}

\paragraph{Results} 
\fref{fig:mnli_temp1} shows explanations and accuracy for instances with template \textit{The [Noun A] [Verb] the [Noun B] $\xrightarrow{?}$ The [Noun B] [Verb] the [Noun A]}, for three studied models, respectively. In the original model, we observe that the overlapping nouns exert large positive attributions ($\phi(x)$), driving the model towards the wrong direction of predicting entailment. Meanwhile, the small negative attributions from the order features ($\phi(z)$) are insufficient to counteract $\phi(x)$. In $M_{aug}$ which scores $100\%$ on this template, we observe that $\phi(x)$ shifts to negative, while $\phi(z)$ remains the same: the model pays no more attention to word orders compared to $M_{ori}$, other than simply reducing the correlations between overlapping words and entailment by shifting the decision boundary. In contrast, $M_{for}$ produces more conceptually sound explanations: $\phi(z)$ almost doubles, while $\phi(x)$ is still positive, albeit smaller in magnitude. However, \sv{} shows no such distinctions: both $M_{aug}$ and $M_{for}$ produces identical explanations, which may be misconstrued as both models successfully overcoming the spurious correlations. 

We demonstrate more general quantitative result in \tref{tab: hans}, by computing the sum of $\phi(x)$ and $\phi(z)$ averaged across all non-entailment-labeled templates: while both $M_{for}$ and $M_{aug}$ suppress the effect of the overlapping words, $M_{for}$ sees a sizable~(15\%) decrease of $\phi(z)$ in contrast to $M_{aug}$, which even increases a little.  More result are included in Appendix~\ref{appendix:hans}.

\textbf{In contrast to standard \sv{} explanations, \osv{} shows that the more conceptually sound model $M_{for}$ recognizes the difference in word orders between the premise and hypothesis as an indicator of non-entailment, while shifting of decision boundary without the true learning of order in $M_{aug}$ may lead to overfitting~\cite{lu2018gender, jha2020does}.} To test this hypothesis, we create a simple ``adversarial'' dataset, HANS*,  by replacing the hypothesis in HANS with the same sentences as the premise, effectively creating instances such as \textit{The doctors visited the lawyers. $\xrightarrow{?}$ The doctors visited the lawyers.}, for which the ground truth is always entailment. According to \tref{tab: hans}, $M_{ori}$ scores $100\%$  and $M_{for}$ maintains a high accuracy of $94\%$ on HANS*, while $M_{aug}$ scores only $64\%$, corroborating the overfitting hypothesis. 

\caleb{sharp statements, start with caption}

\begin{figure}[t]
\centering
\subfloat[$\phi^a(x)$]{\label{fig:sst_good_feature}
     \includegraphics[width=.49\columnwidth]{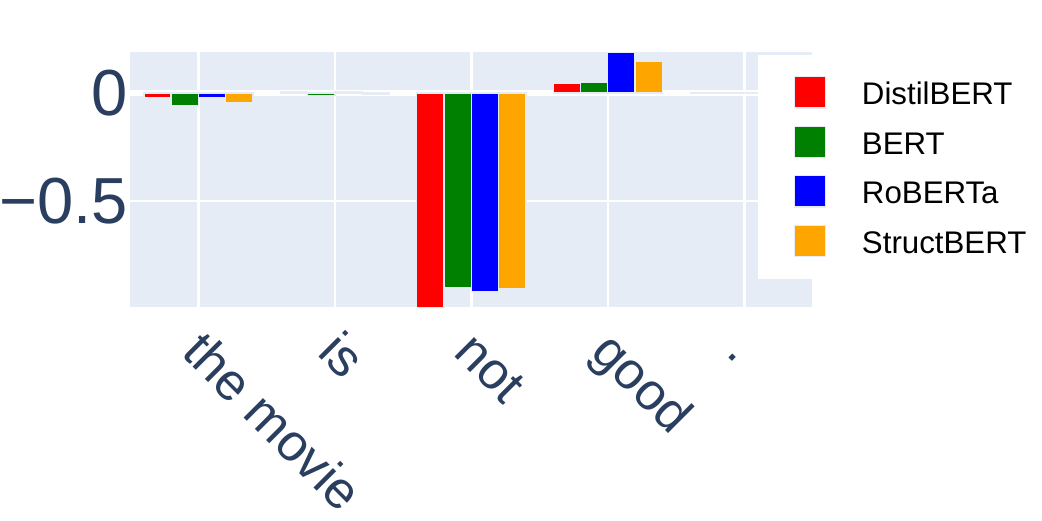}}
\subfloat[$\phi^a(z)$]{\label{fig:sst_good_order}
     \includegraphics[width=.49\columnwidth]{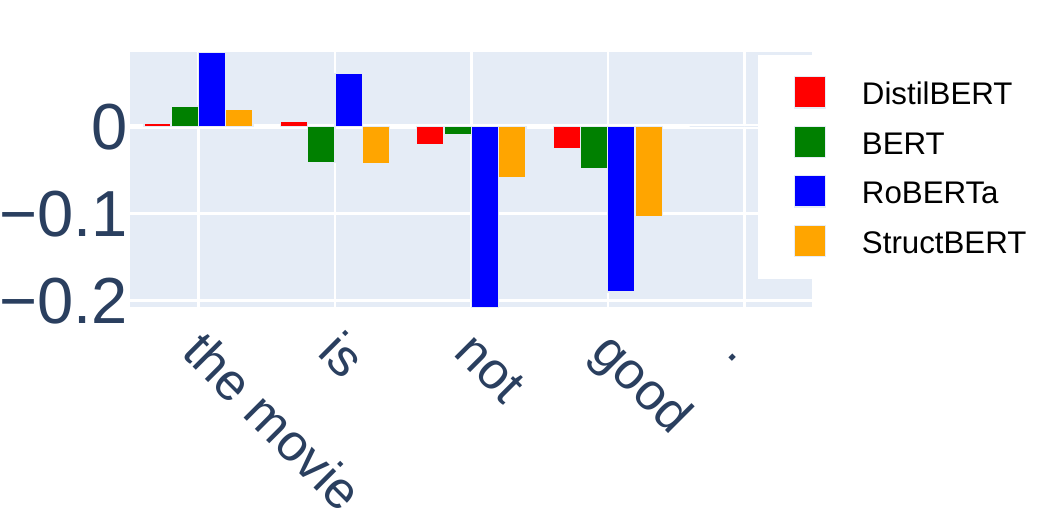}}
\newline
\subfloat[$\phi^a(x)$]{\label{fig:sst_bad_feature}
     \includegraphics[width=.49\columnwidth]{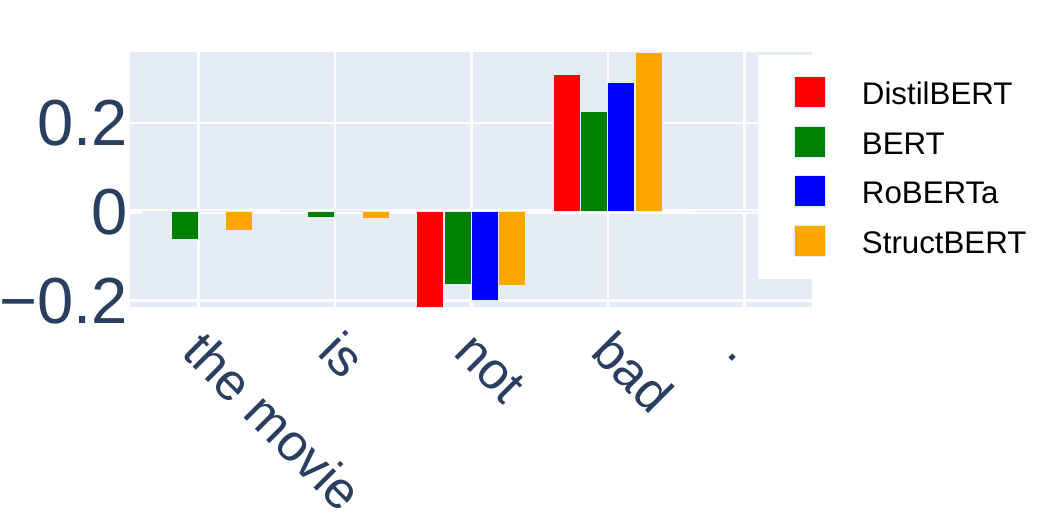}}
\subfloat[$\phi^a(z)$]{\label{fig:sst_bad_order}
     \includegraphics[width=.49\columnwidth]{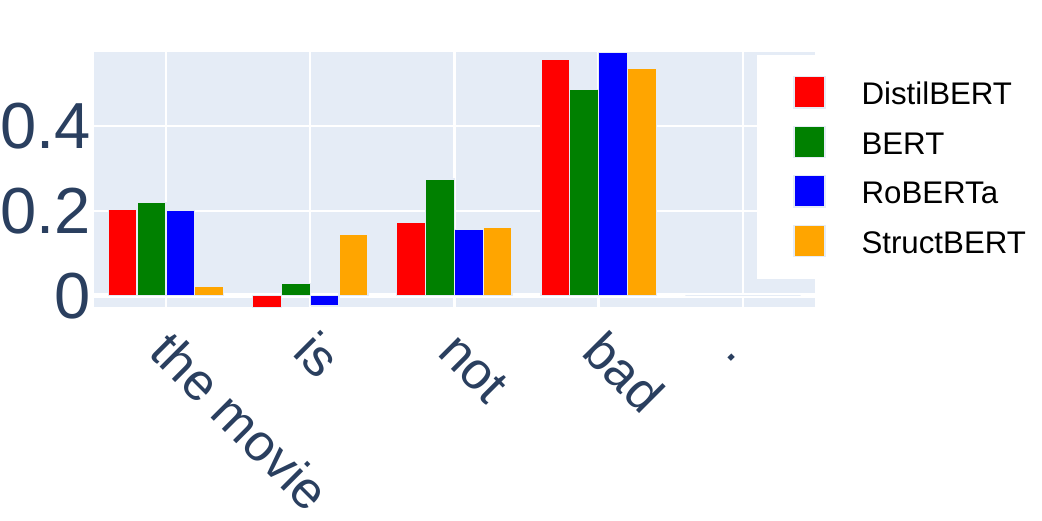}}
\caption{Explanations for two example sentences with negation across four models with $f^y(x) = p_{positive} -p_{negative}$. $\phi^a(x)$ and $\phi^a(z)$ are attributions to occurrence and order features, respectively. \caleb{more captions}}
\label{fig:sst}
\end{figure}

 \begin{table}[t]
\caption{Model performances of sentiment analysis for various datasets.}
\label{tab: sst}
\begin{small}
\begin{center}
\begin{tabular}{lcccc}
\toprule
Dataset & DistilBERT & BERT & RoBERTa & StructBERT \\
\midrule
REG  & 1.0 & 1.0& 1.0 & 1.0\\
NEG   & 0.99 & 1.0& 1.0 & 1.0\\
REG*(+)   & 0.85& 0.85& 1.0&1.0\\
REG*(-)   & 0.96& 0.91& 0.99&1.0\\
REG*,(+)  & 0.96& 0.99& 1.0&1.0\\
REG*,(-)  & 1.0& 1.0& 1.0&1.0\\
REG*B(+)  & 0.99& 1.0& 1.0&1.0\\
REG*B(-)  & 1.0& 1.0& 1.0&1.0\\
\bottomrule
\end{tabular}
\end{center}
\end{small}
\end{table}

\subsubsection{Negation in Sentiment Analysis}\label{sec: sst}

Previous works find that word order hardly matters to Transformer-based sentiment analysis models. Mostly notably,~\cite{pham2020out} discovers that ``more than 60\% of instances can be correctly predicted by the top-1 salient word.''   As a result, we focus on one syntactical relation that is both order-sensitive and essential for classifying sentiment: negation. In particular, we look at the construct of negation cues (ex. \textit{no}, \textit{not}, \textit{never}) preceding an argument word/phrase (\textit{not good} or \textit{not a good movie}). Order is important in these constructs: for instance, \textit{not} should only reverse the sentiment when directly preceding an argument.

\paragraph{Models and Data} We evaluate four models, DistilBERT~\cite{sanh2019distilbert}, BERT-Base~\cite{devlin2018bert}, RoBERTa~\cite{liu2019roberta}, and StructBERT~\cite{wang2019structbert}. All models are trained on 2-class GLUE SST-2 sentiment analysis dataset~\cite{wang2018glue}.  Due to the lack of instances in SST-2 with the targeted negation construct, we construct a simple synthetic dataset to surgically evaluate the model's learning of negation. A total of 3072 sentences such as those in \fref{fig:sst} are generated from the template \textit{[Det(The)] [Noun(movie)] [Verb(is)] [NEG/REG] [Adjective(good)][(Punc).]} where \textit{[REG/NEG]} is evenly distributed between negation cues (ex.\textit{not}) and regular adverbs~(ex.\textit{very}) or blank. The adjectives are evenly distributed between common positive and negative sentiment adjectives(See Appendix~\ref{appendix:sst} for more details). All models perform really well on both the regular (REG) and negated (NEG) half of the dataset, as shown by the first two rows of \tref{tab: sst}.

\begin{figure*}[t]
\subfloat[SVA: across Object Relative Clause]{\label{fig:sva_obj}
     \includegraphics[width=0.8\textwidth,valign=c]{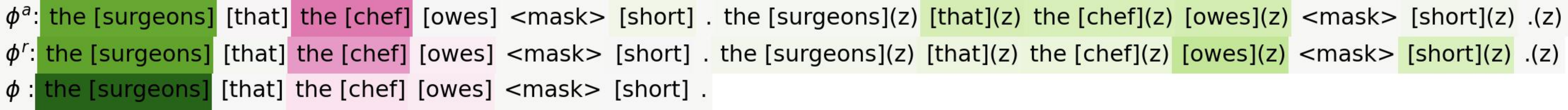}}

\subfloat[SVA: across Subject Relative Clauses]{\label{fig:sva_subj}
     \includegraphics[width=0.8\textwidth,valign=c]{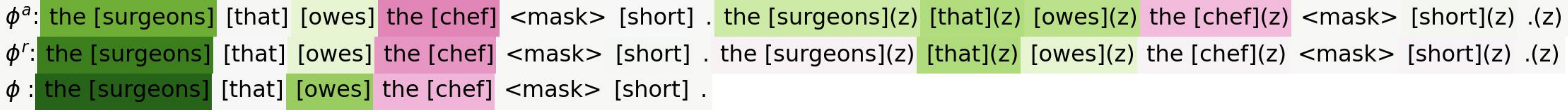}}

\caption{Global explanations for two SVA tasks for DistilBERT. Words in brackets ([]) are one example filling in variable template positions. Features suffixed by (z) are order features. $f^y(x) = p_{correct\_verb} -p_{incorrect\_verb}$. Positive attributions are in green and negative in pink. Darker shade represents larger attribution magnitude.}
\label{fig:sva}
\end{figure*}



\paragraph{Results}
\fref{fig:sst} shows local \osv{}($\phi^a$) with marginal intervention with a single token \textit{[MASK]},  for all four models on two example instances. We observe an interesting distinction between the two:
High $\phi(z)$ compared to $\phi(x)$ on \textit{not} and \textit{bad} across all models in \fref{fig:sst_bad_feature} indicates that it is essential for \textit{not} to precede \textit{bad} to correctly reverse the sentiment. However $\phi(z)$ on \textit{not good}~(\fref{fig:sst_good_feature}), is comparably smaller in magnitude for all models, particularly for DistilBERT and BERT, where the attributions to order features are negligible. This discrepancy is mostly likely caused by the inherent polarity of the word \textit{not}, hence models do not have to encode order for negating positive adjectives while order is essential for negating negative adjectives. \caleb{more results in appendix, add another sentence that elaborates a little bit. really surfacing through the explanations, bring out the differences in terms of how models are processing. bold provide validation for explanations.} Similar polarity of negation cues are also observed in prior works~\cite{lu2021influence, chen2020ls}, while \osv{} provides a more precise analytical approach to surface it. One possible reason for the model-wise difference is that RoBERTa and StructBERT benefit from their improved training procedures over BERT/DistilBERT. StructBERT, for example, is specially designed to be sensitive to word orders during pretraining.  

To demonstrate the potential weakness of such order insensitivity of BERT and DistilBERT, we construct an adversarial dataset (REG*) by appending one of five neutral phrases containing negation words(\textit{not as expected}, \textit{not like the other}, \textit{not gonna lie}, \textit{never gonna lie}, \textit{never as expected}), to the end of sentences of REG, effectively creating sentences such as \textit{The movie is good not gonna lie}, which happens to contain the flipped negation construct(\textit{good not}) and \textit{not} in this context should not convey any sentiment. This dataset of size 7680 is evenly distributed between negative and positive sentiments. The result is shown in \tref{tab: sst}. Mirroring the difference in order-sensitivity shown by $\phi(z)$, DistilBERT and BERT both suffer a 15\% decrease for sentences with positive sentiments~(REG*(+)) while the accuracy is either unchanged for RoBERTa and StructBERT, or decreases slightly for sentences with negative sentiments~(REG*(-)) for all models. To further ensure the drop is not merely caused by the inherent negative sentiment of the word \textit{not} (apart from the same $\phi(x)$ for \textit{not} in \fref{fig:sst_good_feature}), we test two alternative datasets: (1) REG*, inserts a comma between the appended phrase and the original sentences, ex.: \textit{The movie is good, not gonna lie.} (2) REG*B appends the phrases to the beginning of the sentences instead.  Interestingly, all four models are robust to both alternatives. This distinction implies that the models are capable of disregarding the added phrases if negation cues are not directly proximate to the adjective. \textbf{\osv{}, unlike \sv{}, effectively surfaces local order-insensitivity in the negation construct, which is validated by BERT \& DistilBERT's failure on REG*(+). }

\begin{table}[t]
\caption{Absolute-relative order discrepancy and Model performances for SVA. }
\label{tab: sva}
\begin{small}
\begin{center}
\begin{tabular}{p{1.2cm}p{.55cm}p{1.1cm}p{.4cm}|p{.55cm}p{1.1cm}p{.4cm}}
\toprule
&  \multicolumn{3}{c|}{SVA-Obj} & \multicolumn{3}{c}{SVA-Subj} \\
& $\Delta\phi^{z}_{su}$  & Acc.* & Acc.  & $\Delta\phi^{z}_{su}$  & Acc.* & Acc. \\
\midrule
DistilBERT  &.04&.86($\downarrow$.05)  &.91 &.12 &.79($\downarrow$.14)  &.92 \\
BERT-B &-.03 &.93($\downarrow$.00)  &.93 &.11  &.94($\downarrow$.04)  &.99\\
BERT-L   &-.05 &.81($\downarrow$.00)  &.81 &.09&.92($\downarrow$.04)  &.96\\
RoBERTa     &-.04 &.71($\downarrow$.00)  &.71 &.09 &.64($\downarrow$.08)  &.72\\

\bottomrule
\end{tabular}
\end{center}
\end{small}
\vskip -0.1in
\end{table}

\subsubsection{Subject-Verb Agreement(SVA)}\label{sec: sva}

While fine-tuned classification models may not learn word order because they don't have to, a pretrained language model should, in theory, be more sensitive to order, especially when evaluated on syntactic tasks such as Subject-Verb Agreement. The task of SVA evaluates whether a language model prefers the correct verb form to match with the subject. For example, a pretrained masked language model should assign higher probability to \textit{is} over \textit{are} for the sentence \textit{The cat <mask> cute}. 

How language models learn SVA is widely studied in prior works~\cite{linzen2016assessing, DBLP:conf/acl/LuMLFD20, wei2021frequency}. \caleb{more} In this section, we focus on two complicated long-range formulations of SVA where Transformers excel over simple RNN models~\cite{goldberg2019assessing}: SVA across subject relative clauses (\textit{SVA-subj}) and SVA across object relative clauses (\textit{SVA-obj}). 
For example, test instances for \textit{SVA-obj} are generated from the template: \textit{the [Subject] that the [Attractor] [Verb] <mask?(is/are)> [Adjective].}(ex.\fref{fig:sva_obj}). A language model needs to correctly parse the sentence to identify the true subject. In particular, when \textit{Attractor} is of an opposite number to the \textit{Subject}, a model that fails to parse the sentence will be distracted by the more proximate attractor. Prior works find that Transformer models succeed in long-range SVA and overcome such distraction because they learn composite syntactic structures~\cite{jawahar2019does, hewitt2019structural}. 

\paragraph{Models and Data}  We use dataset from~\cite{marvin2018targeted} and the cloze-style set up used in~\cite{goldberg2019assessing}. Specifically, we look at sentences with oppositely numbered attractors. We evaluate four pretrained langauge models: DistilBERT~\cite{sanh2019distilbert}, BERT-Base \& BERT-Large~\cite{devlin2018bert}, and RoBERTa~\cite{liu2019roberta}. The accuracy of each task is found in \tref{tab: sva}, where all models perform quite well except for RoBERTa, which still scores more than 70\%. 
\paragraph{Result}
\fref{fig:sva} show the global explanations for \textit{SVA-Obj} and \textit{SVA-Subj} with $f^y(x) = p_{correct\_verb} - p_{incorrect\_verb}$ and the same $g$ as~\sref{sec: hans}. We observe that the subject word exerts positive attribution on the correct choice of the verb, and the intervening noun (attractor) exerts negative attribution.  In contrast to fine-tuned classification models such as NLI and SST, attributions to order features are high: they are indeed more important for syntactical tasks.  Zooming in on the explanations, we obtain two other findings:

First, we observe that the relativizer word \textit{that} is important as an order feature, but not as an occurrence feature. In other words, in order to predict the correct verb form, \textit{that} itself may not contain the number signal, nevertheless it is regarded by the language model as an essential syntactic boundary.

Second, there is a big difference~($\Delta\phi^{z}_{su}$) between $\phi^a$ and $\phi^r$ for the subject word in \textit{SVA-Subj.}, but not for \textit{SVA-Obj.}, i.e, the model seems to rely on absolute position of the subject word in \textit{SVA-Obj.} much more than the relative position of the subject word. $\phi^a$ of the attractor is also much more negative than $\phi^r$, indicating the model is distracted by the proximity of the attractor words (along with their occurrences). These distinctions point to a hypothesis that even though they contain exactly the same words, the model parses \textit{SVA-Obj.} better than \textit{SVA-Subj.}, most likely due to the proximity of the attractor with \textit{<mask>} in \textit{SVA-Subj}. \textbf{The higher accuracies of \textit{SVA-Obj.} may be attributed to the models' reliance on the absolute position of the subject words in the beginning of sentences, surfaced by the two modes of intervention on order features($q^a$ \& $q^r$ in \sref{sec: replacing}). Without attributing to order features, however, \sv{} cannot uncover this subtle discrepancy.} \caleb{more results in appendix}

To test this hypothesis, we create a simple adversarial set by prepending each of 23 punctuation and symbols\footnote{\lstinline{!"#&\'*+,-./:<=>?@^_|~;<unk>}} such as \textit{``.''}  or \texttt{<unk>} to the beginning of sentences and see if such shifting of the subject from its absolute position impact the predictions. In \tref{tab: sva}, Acc* shows all models suffer more from this adversarial perturbation for \textit{SVA-Subj} than \textit{SVA-Obj}, mirroring the distinction from $\Delta\phi^{z}_{su}$. This indicates that the models' learning of long-range SVA is not always conceptually sound, as is in the case of $\textit{SVA-Subj.}$. 
\section{Related Work}\label{sec:related}
Recent works ~\cite{pham2020out, sinha2021masked, clouatre2021demystifying, alleman2021syntactic, sinha2021unnatural, gupta2021bert} show NLP models' insensitivity to word orders. Our work corroborates such insensitivity by showing often insignificant attributions to order features in classification models. Meanwhile, a systematic explanation device such as \osv{} offers more precise and measurable attribution to the importance of order. 

 Shapley Values~\cite{shapley1953contributions} is widely adopted as an model-agnostic explanation method~\cite{strumbelj2010efficient, datta2016algorithmic,lundberg2017unified, sundararajan2020many, covert2020understanding} and also applied to explain text classification models~\cite{chen2020ls, zhang2021building, chen2018shapley}. However, the explanation power of these methods is limited by treating text data as tabular data. \osv{}, as demonstrated by numerous examples, allows us inspect the effect of ``word occurrences'' along side the effect of ``word orders''. As order sensitivity is an essential metric of conceptual soundness, explanations incorporating order are more faithful in reflecting the true behavior of models. 

We believe that explanations should not only function as a verification for plausibility~\cite{jacovi2020towards}, but also as a tool for diagnosing conceptual soundness. \textbf{Instead of \emph{confirming} the human understanding or the correct linguistic rules, explanations are equally, if not more, insightful when \emph{deviating} from them.}  One popular approach to surface such deviation is adversarial examples~\cite{zhang2020adversarial}, and many algorithms for finding them in NLP tasks~\cite{cheng2020seq2sick, ebrahimi2017hotflip, wallace2019universal} use gradient-based attributions such as saliency maps to guide adversarial perturbations. However, the use of orderless explanations is limited to local perturbations, such as swapping words into their synonyms. To allow for broader and non-local testing of model robustness, challenge datasets~\cite{belinkov2017neural,ribeiro2020beyond, wu2021polyjuice, mccoy2019right} are constructed to stress test NLP models. These tests are becoming more crucial in designing and evaluating the conceptual soundness and robustness of NLP models beyond benchmark performances. However, rarely is the creation of those datasets guided by explanations methods, making it difficult to systematically and deductively discover non-local adversarial examples. In this paper, we show an initial attempt to bridge the gap between the two: we show that \osv{} not only informs on \emph{which}, but \emph{how} adversarial perturbations break models, enabling a deeper understanding of model weaknesses than orderless explanations. More broadly, the findings in \sref{sec:nlp} echos~\cite{lovering2020predicting} and~\cite{mangalam2019deep}, which discover that models learn ``easier features''(e.g.\ occurrence features in HANS, \textit{not} in SA and absolute order features in SVA) before learning ``harder features'' that may actually be the conceptually sound ones.



\section{Conclusion}\sm{(.25 page)}
We propose \osv{} for explaining word orders of NLP models. We show how \osv{} is an essential extension of Shapley values and introduce two mechanisms for intervening on word orders. We highlight how \osv{} can precisely pinpoint if, where and how a model uses word order. Using adversarial examples guided by \osv{}, we demonstrate how order insensitivity, harmless as it seems, result in conceptually unsound models.
\paragraph{Acknowledgement}
This work was developed with the support of NSF grant CNS-1704845. The U.S. Government is authorized
to reproduce and distribute reprints for Governmental purposes not
withstanding any copyright notation thereon. The views, opinions,
and/or findings expressed are those of the author(s) and should not
be interpreted as representing the National Science
Foundation or the U.S. Government. We gratefully acknowledge the support of NVIDIA Corporation with the donation of the Titan V GPU used for this work.
\newpage
\bibliographystyle{IEEEtran}
\bibliography{main}
\newpage
\appendix
\section{Code and Data}
Code is included in the supplementary material. Data for SA is introduced in Appendix~\ref{appendix:sst} while all other data in \sref{sec:nlp} are publicly available as benchmark datasets. The code for synthesizing the data in \sref{sec: toy} is included in the code files. 

\section{Proof of Remark~\ref{rem: remark1} and Axiomatic Interpretations of \osv{}}\label{appendix:method}
\subsection{Proof of remark \ref{rem: remark1}}
An alternative definition\cite{datta2016algorithmic} for $\phi^z_{f^y}(i)$ is:
\begin{align}
 \forall x_i \in x,
    \phi^z_{f^y}(x_i)&=\frac{1}{(2n)!}\sum_{\sigma \subseteq \Pi(N')} m_i(\sigma) \notag \\
                &=\frac{1}{(2n)!} \cdot \frac{(2n)!}{n!} \sum_{\sigma\subseteq \Pi(N)} m_i(\sigma) \notag \\
                &=\frac{1}{n!}\sum_{\sigma\subseteq \Pi(N)} m_i(S) \notag \\
                &=\phi_{f^y}(x_i) \notag \\
where: m_{i}(\sigma)&=m\left(P_{i}(\sigma)\right) \notag \\
P_{i}(\sigma)&=\{j \mid \sigma_j<\sigma_i\} \notag
\end{align}
$P_{i}$ is the set of $i$’s predecessors in $\sigma$.
$\forall z_i \in z$, according to the dummy axiom, since $z$ is omnipresent, $f^y(S\cup i) = f^y(S) $, so $\phi^z_{f^y}(z_i) = 0, \forall z_i \in z$. 
\subsection{Axioms of Shapley Values}
The axiomatic properties of Shapley values are first introduced in~\cite{shapley1953contributions} and extensively discussed in the context of machine learning in~\cite{lundberg2017unified} and~\cite{datta2016algorithmic}. Shapley values defined in~\eref{eq:shapley} is the only value satisfying all following axioms. The names in parenthesis are alternative names used in literature. 

\paragraph{Symmetry Axiom}
$i,j\in N$ are \emph{symmetric} if $v(S\cup i) = v(S\cup j)$ for all $S\subseteq N \backslash {i,j}$. A value $\phi$ satisfies symmetry if $\phi(i) = \phi(j)$ whenever $i$ and $j$ are symmetric. 

Extending to order features, if two order features are symmetric, meaning that their positions are always interchangeable, then their attributions should equal each other. Though theoretically possible, order features are unlikely to be symmetric with occurrence features.  
\paragraph{Dummy Axiom(Null Effects/Missingness)} A player is a dummy or null player if $v(S\cup i) = v(S)$ for all $S\subseteq N$. A value $\phi$ satisfies the dummy axiom if $\phi(i)=0$ whenever $i$ is a dummy or null player. 

An order feature is a dummy: (1) if the model doesn't care where it is in a model, for example a BoW model; (2) if the order feature is omnipresent as in Remark~\ref{rem: remark1}.
\paragraph{Completeness Axiom (Efficiency or Local Accuracy)} 
A value $\phi$ satisfies completeness if $\sum_i^{N} \phi(i) = v(N)$. This axiom is important as it makes Shapley value an \emph{attribution value} in that it \emph{attributes} and \emph{allocates} the output of the function to individual inputs. 

\paragraph{Monoticity Axiom (Consistency)}
A value $\phi$ satisfies monoticity if $m(S, v_1) \geq m(S, v_2)$ for all $S$ implies that $\phi_{v_1}(i) \geq \phi_{v_2}(i)$.
The theorem of monoticity states that if one features' contribution is greater for one model than another model regardless of all other inputs, that input's attribution should also be higher. It can be extended directly to order features. 

\section{Implementation details}\label{appendix:impl}

To make sure that the global explanations truly represent the global contribution of each feature towards the prediction, we use a convergence factor of $t = 0.005$ for all global explanations, meaning the variance of the Shapley value is less than $0.5\%$ of the difference between the maximum and minimum $\phi(i)$ for all $i$ (a looser $t = 0.01$ is sufficient for convergence according to~\cite{covert2020understanding}). For each $S$, we use a sample size of 4 for $q$ and a sample size of 5 for $g$ when $g$ samples from templates  (total of 20 intervened sentences per instance per $S$). We find this setting generates stable explanations for the synthetic data (low variance in \tref{tab: toy-quan}). Moreover, since the stopping condition is based on $t$ and the algorithm for computing global explanations may iterate through the whole data multiple times to reach that convergence, the actual sample size of $q$ and $g$ in principle should not matter. 

For StructBERT~\cite{wang2019structbert} and $M_{for}$of~\sref{sec: hans}, we use the implementation from the official repositories. For other models including $M_{aug}$ of~\sref{sec: hans}, we either use models available in the Huggingface~\cite{wolf2019huggingface} repository or fine-tuned each model with 3 epochs using the default parameters of HuggingFace Trainer class. We do not make extensive efforts to tune the hyperparameters since the goal of this paper is not to use the best models but to show the utility of an explanation device.

\tref{tab: toy-compute} shows the average number of evaluations (number of intervened instances) per instance for computing global explanations shown in \fref{fig:toy_lstm} and \ref{fig:toy_transformer}. For sequences of length 8, $\phi^z$ needs less than 8 times more evaluation for convergence, corroborating Theorem 2 of~\cite{covert2020understanding}. 

Depending on the length of the sequence, GPU run time on a  for computing~\osv{} varies. In this paper, all experiments finish with a reasonable amount of time even with the tight convergence setting. For example, on average, each template in~\sref{sec: hans} takes around 5 min. 
\begin{table}[h]
\caption{Average (across 25 seeds) number of evaluations per instance for synthetic experiments}
\label{tab: toy-compute}
\begin{small}
\begin{center}
\begin{tabular}{lll}
\toprule
& LSTM & Transformer \\
\midrule
$\phi^a$  & 2425 & 2299\\
$\phi^r$ & 2485 & 2445\\
$\phi$  & 330 & 331\\

\bottomrule
\end{tabular}
\end{center}
\end{small}
\vskip -0.1in
\end{table}

\section{Additional Results for \sref{sec: toy}}\label{appendix:toy}
\begin{figure}[ht]
\centering
\subfloat[Model 1]{\label{fig:toy_transformer_1}
     \includegraphics[width=.85\columnwidth]{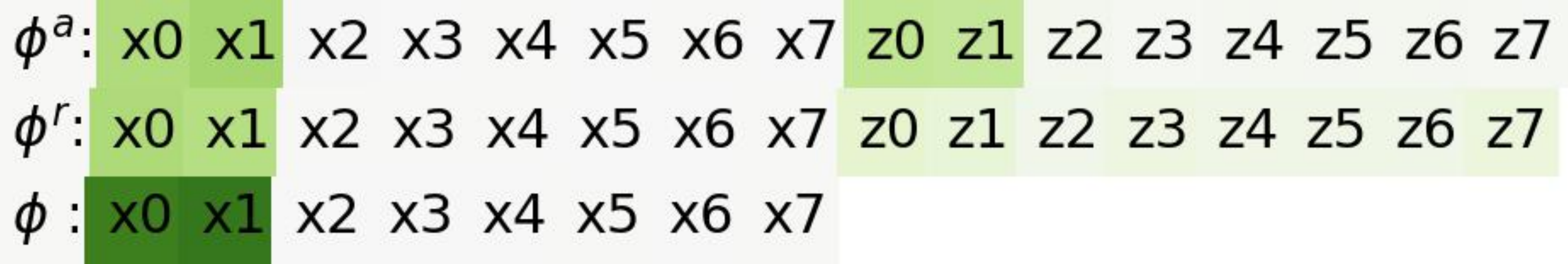}}

\subfloat[Model 2]{\label{fig:toy_transformer_2}
     \includegraphics[width=.85\columnwidth]{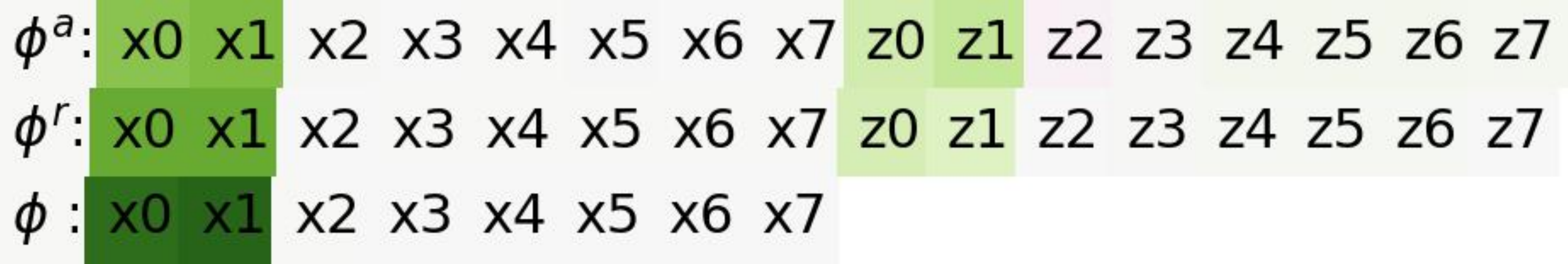}}

\subfloat[Model 3]{\label{fig:toy_transformer_3}
     \includegraphics[width=.85\columnwidth]{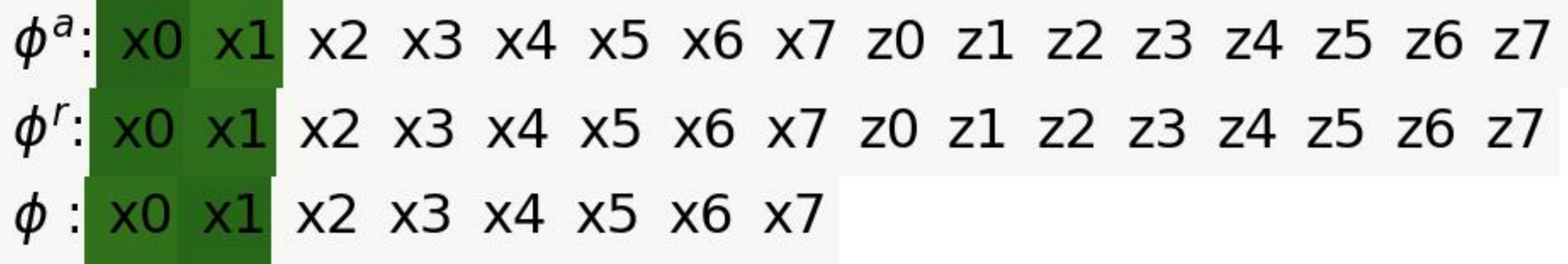}}
\caption{Explanations($\phi^a$, $\phi^r$ and $\phi$) of $\seq{W_1}$ by three Transformer models solving tasks in \tref{tab: toy-desc-transformer}}
\label{fig:toy_transformer}
\end{figure}

For IG, we use 1000 as the number of steps for approximation(the authors of IG recommends 20 to 1000 steps). The model performance and explanation visualizations for transformer models are included in~\tref{tab: toy-desc-transformer} and~\fref{fig:toy_transformer}.
\begin{table}[h]
\caption{Tasks Description and Model performances for Synthetic Data Experiments for transformer models. Acc. are test accuracies for each model over 5 random seeds, Acc-1 are accuracies for $\seq{W_1}$.}
\label{tab: toy-desc-transformer}
\begin{small}

\begin{center}
\begin{tabular}{lcccr}
\toprule
Ind. & Condition for y=1 & Acc. & Acc-1. \\
\midrule
1    & Begins with duplicate & 1.00$\pm$ .00& 1.00$\pm$ .00\\
2   & Adjacent duplicate & 0.98$\pm$ .00&0.99$\pm$ .01 \\
3    & Any duplicate &1.00$\pm$ .00& 1.00$\pm$ .00 \\

\bottomrule
\end{tabular}
\end{center}
\end{small}
\end{table}

\section{Additional Results for HANS~(\sref{sec: hans})}\label{appendix:hans}
We compute the global explanations for all templates in~\cite{mccoy2019right}, where half of the templates is labeled as entailment. For those templates, word overlapping heuristics is actually the correct heuristic for predicting entailment. \tref{tab: hans-entailment} shows the same stats as \tref{tab: hans}, but for entailment-labeled templates. As we can see, $M_{for}$ is actually less order-sensitive than $M_{aug}$. Ideally, a conceptually sound model should learn both word overlaps and word orders; nevertheless this discrepancy can also be explained by overfitting and mere shifting of decision boundary:  $M_{aug}$ only learn that word overlap at the corresponding positions does not equate to entailment, making the model more order-sensitive for entailment-labeled sentences. While $M_{for}$ also rely on word-overlapping features much more than order, but possibly in the same way as $M_{ori}$. 
\begin{table}[h]
\caption{Global Statistics across entailment-labeled templates}
\label{tab: hans-entailment}
\begin{small}
\begin{center}
\begin{tabular}{lcccc}

\toprule
 & $\sum_i\phi(x_i)$   & $\sum_i\phi(z_i)$  \\
\midrule
$M_{ori}$ & 1.26 & -0.37 \\
$M_{aug}$  &0.85($\downarrow$0.41)  &-0.09($\uparrow$0.28) \\
$M_{for}$  &0.89($\downarrow$0.37)  &-0.22($\uparrow$0.15) \\

\bottomrule
\end{tabular}
\end{center}
\end{small}
\end{table}
\section{Additional Details for SA~(\sref{sec: sst})}\label{appendix:sst}
\subsection{Templates for NEG/REG}
\textit{[Det(The)] [Noun(movie)] [Verb(is)] [NEG/REG] [Adjective(good)][(Punc).]}
\begin{itemize}
    \item Det: 'the', 'that', 'this', 'a'
    \item Noun: 'film', 'movie', 'work', 'picture'
    \item Verb: 'is', 'was'
    \item NEG: 'not', 'never', 'not that', 'never that'
    \item REG: ''(blank), 'very', 'pretty', 'quite'
    \item Adjective: (+): 'good', 'amazing', 'great'
    \item Adjective: (-): 'boring', 'bad', 'disappointing'
    \item Punctuations: '.', '!'
\end{itemize}
\subsection{Aggregated Results for~\fref{fig:sst}}
\begin{figure}[h]
\centering
\subfloat[$\phi^a(x)$]{\label{fig:sst_good_feature_global}
     \includegraphics[width=.49\columnwidth]{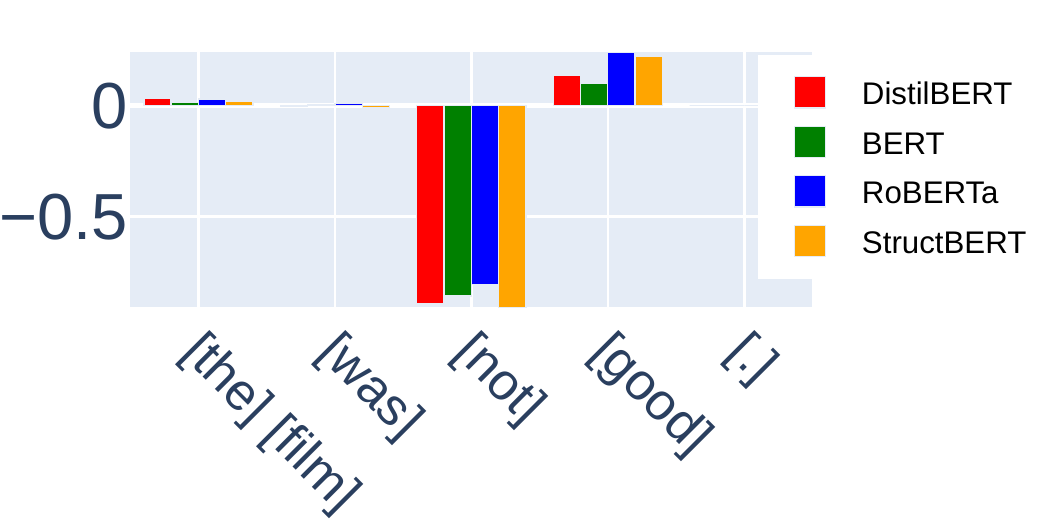}}
\subfloat[$\phi^a(z)$]{\label{fig:sst_good_order_global}
     \includegraphics[width=.49\columnwidth]{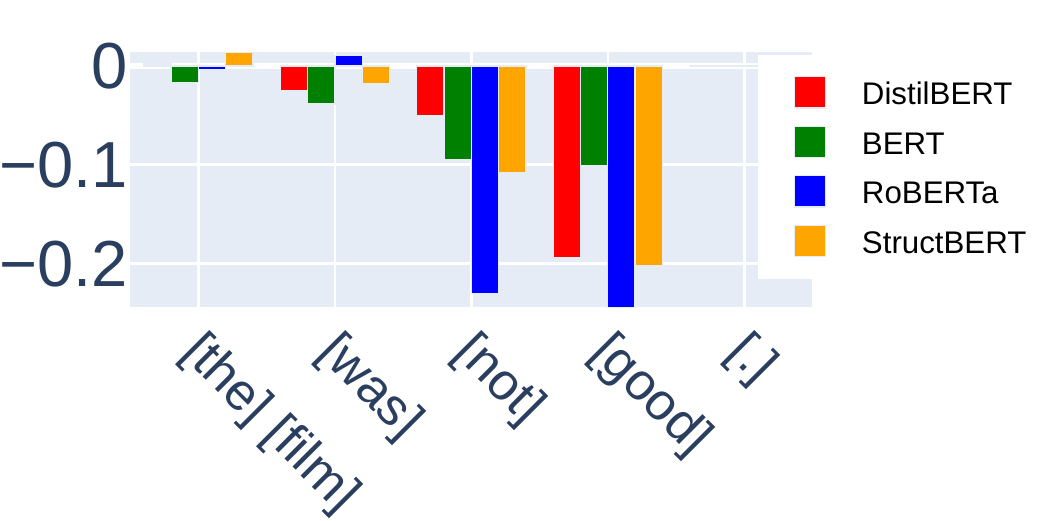}}
\newline
\subfloat[$\phi^a(x)$]{\label{fig:sst_bad_feature_global}
     \includegraphics[width=.49\columnwidth]{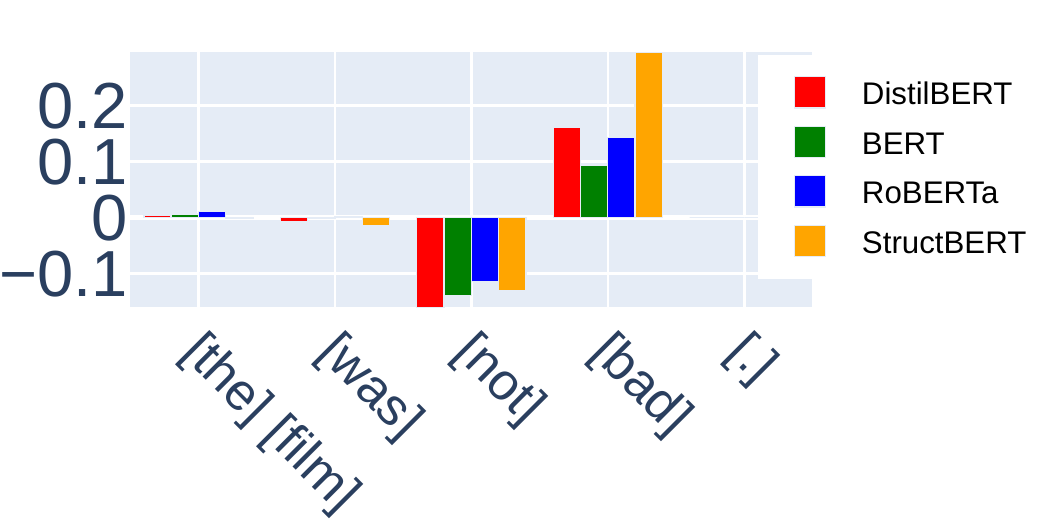}}
\subfloat[$\phi^a(z)$]{\label{fig:sst_bad_order_global}
     \includegraphics[width=.49\columnwidth]{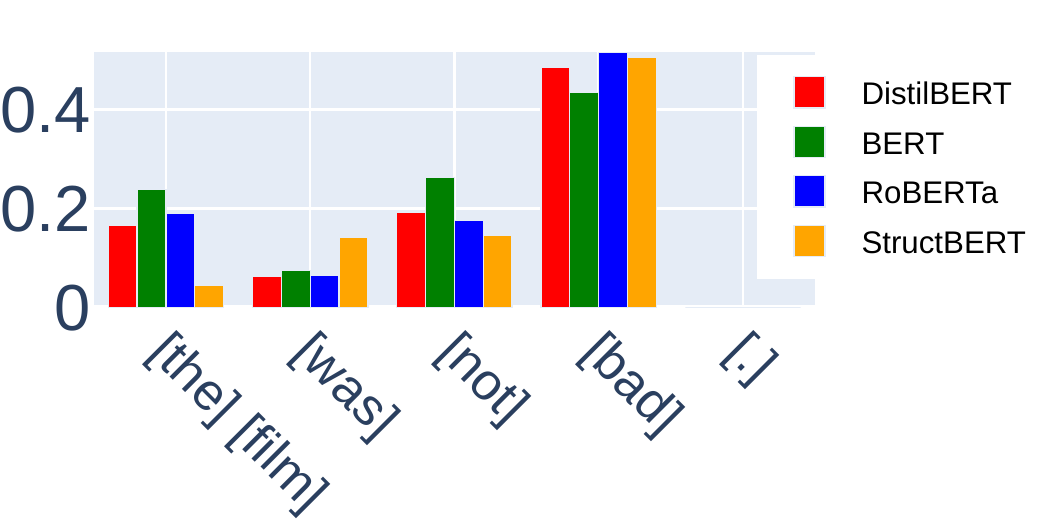}}
\caption{Global explanations for NEG(+) and NEG(-), annotations follow~\fref{fig:sst}}
\label{fig:sst_global}
\end{figure}

\begin{figure}[h]
\centering
\subfloat[$\phi^a$]{\label{fig:sst_good_very}
     \includegraphics[width=.9\columnwidth]{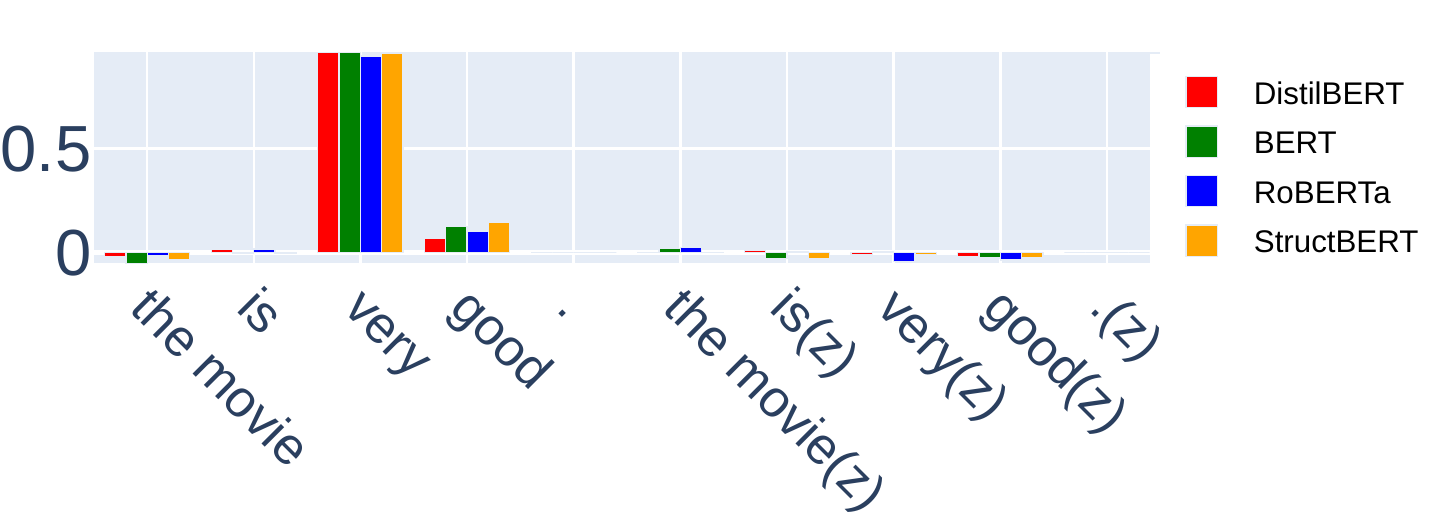}}
\newline
\subfloat[$\phi^a$]{\label{fig:sst_bad_very}
     \includegraphics[width=.9\columnwidth]{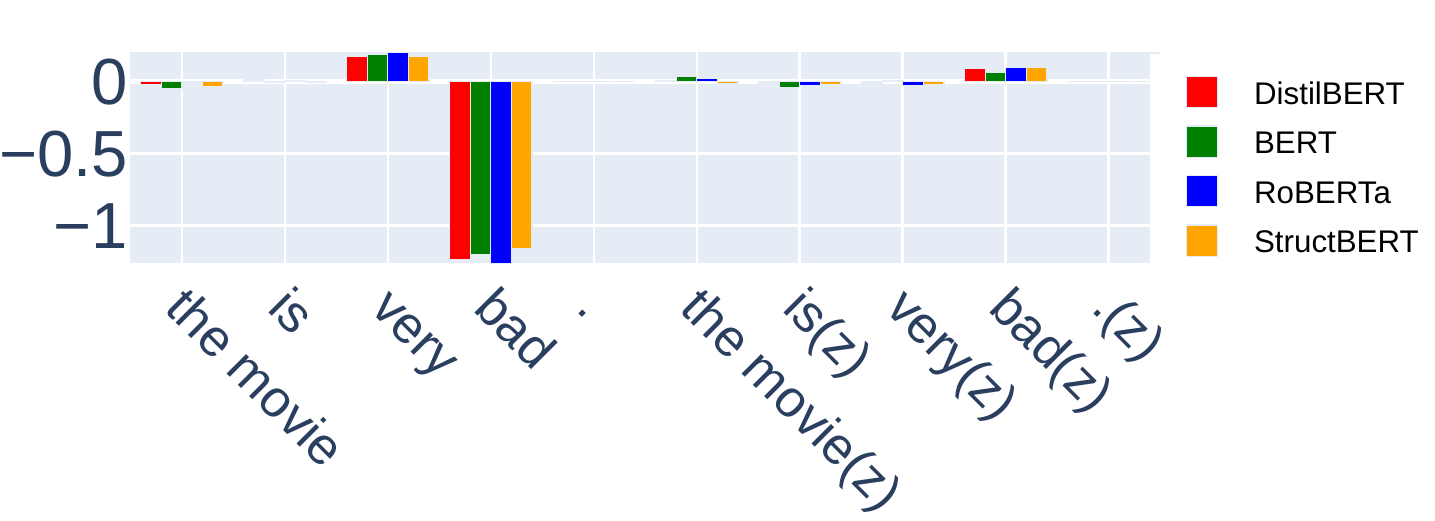}}

\caption{Local explanations for two examples in REG}
\label{fig:sst_very}
\end{figure}

We compute the aggregated global explanations for all instances of NEG(+) and NEG(-), illustrated in \fref{fig:sst_global}, to make sure the trend shown in \fref{fig:sst} represent the general trend. In \fref{fig:sst_very}, we show local explanations for a sentence in REG with no special construct like negation, and as expected, word orders do not matter at all. 

\end{document}